\documentclass{bmvc2k}

\title{InceptionMamba: Efficient Multi-Stage Feature Enhancement with Selective State Space Model for Microscopic Medical Image Segmentation}

\addauthor{Daniya Najiha Abdul Kareem}{daniya.kareem@mbzuai.ac.ae}{1}
\addauthor{Abdul Hannan}{hannan.ansari97@gmail.com}{2}
\addauthor{Mubashir Noman}{mubashir.noman@mbzuai.ac.ae}{1}
\addauthor{Jean Lahoud}{jean.lahoud@mbzuai.ac.ae}{1}
\addauthor{Mustansar Fiaz}{mustansar.fiaz@gmail.com}{3}
\addauthor{Hisham Cholakkal}{hisham.cholakkal@mbzuai.ac.ae}{1}

\addinstitution{
 Mohamed Bin Zayed\\
 University of AI\\
 AbuDhabi, UAE
}
\addinstitution{
 University of Trento\\
 Italy
}
\addinstitution{
IBM Research\\
AbuDhabi, UAE
}

\runninghead{Student, Prof, Collaborator}{BMVC Author Guidelines}

\usepackage{wrapfig}
\usepackage{amssymb}
\usepackage{multirow}

\begin{document}

\maketitle
\begin{abstract}
Accurate microscopic medical image segmentation plays a crucial role in diagnosing various cancerous cells and identifying tumors. Driven by advancements in deep learning, convolutional neural networks (CNNs) and transformer-based models have been extensively studied to enhance receptive fields and improve medical image segmentation task. However, they often struggle to capture complex cellular and tissue structures in challenging scenarios such as background clutter and object overlap.   
Moreover, their reliance on the availability of large datasets for improved performance, along with the high computational cost, limit their practicality.
To address these issues, we propose an efficient framework for the segmentation task, named InceptionMamba, which encodes multi-stage rich features and offers both performance and computational efficiency.  
Specifically, we exploit semantic cues to capture both low-frequency and high-frequency regions to enrich the multi-stage features to handle the blurred region boundaries (e.g., cell boundaries).  
These enriched features are input to a hybrid model that combines an Inception depth-wise convolution \cite{szegedy2015going} with a Mamba block \cite{gu2023mamba}, to maintain high efficiency and capture inherent variations in the scales and shapes of the regions of interest.
These enriched features along with low-resolution features are fused to get the final segmentation mask. Our model achieves state-of-the-art performance on two challenging microscopic segmentation datasets (SegPC21 and GlaS) and two skin lesion segmentation datasets (ISIC2017 and ISIC2018), while reducing computational cost by about 5 times compared to the previous best performing method.
\end{abstract}

\section{Introduction} \label{sec:intro}
Microscopic image segmentation is an essential task in the medical domain as it assists specialists in diagnosing various diseases accurately and thoroughly. 
Additionally, it provides valuable information in other medical applications such as drug discovery and cellular analysis \cite{leygeber2019analyzing}. 
However, medical image segmentation is a challenging task due to the presence of irregularly shaped objects consisting of indistinct boundaries having various sizes, especially for the microscopic images \cite{fiaz2023sa2}. The problem becomes further challenging in the presence of complex backgrounds and clutter. 
For instance, the texture, color, and shape of cells and tissues in microscopic images from different organs present significant difficulties for accurate segmentation. 
\begin{wrapfigure}{r}{0.55\textwidth}
\vspace{-0.25cm}
  \centering
    \includegraphics[width=1.0\linewidth]
 {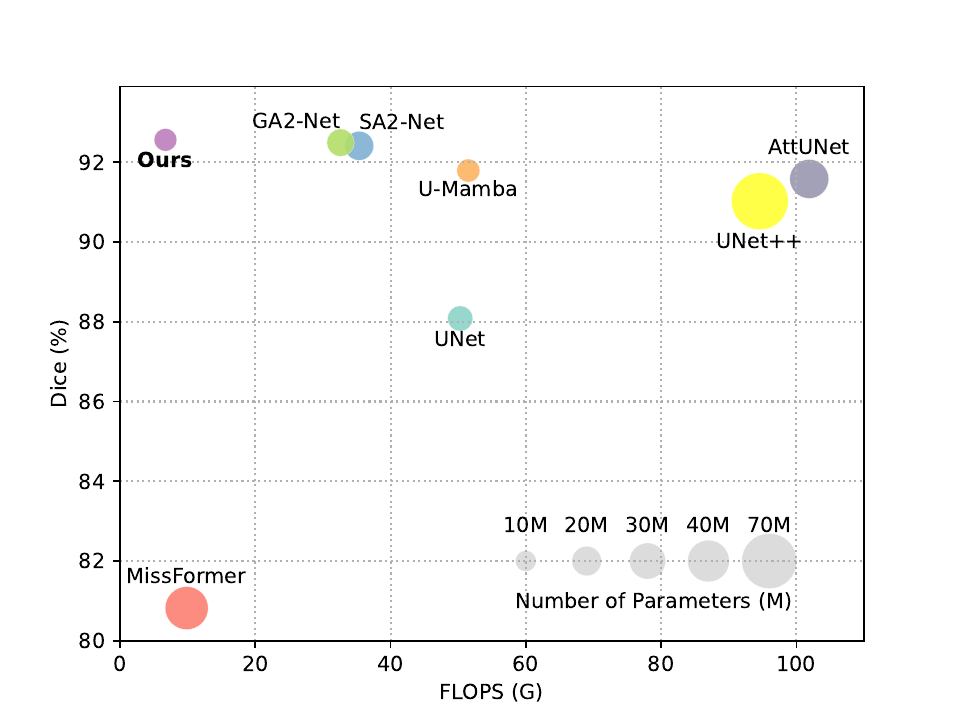} 
 \vspace{-0.5cm}
 \caption{Comparison of our method with state-of-the-art models based on Dice score, parameters, and GFLOPs on the SegPC21 dataset. Our method performs favorably compared to the state-of-the-art GA2-Net while requiring $4.85\times$ fewer GFLOPs and $1.45\times$ fewer parameters.}
    \label{fig:chart}
    \vspace{-0.1cm}
\end{wrapfigure}
Earlier segmentation methods in the medical domain were composed of convolutional neural networks (CNNs) in an encoder-decoder fashion. UNet \cite{ronneberger2015u}, introduced a decade ago gained popularity for the medical segmentation task. 
Subsequently, many researchers adapted this architecture due to its promising performance, leveraging dense skip connections and multi-scale feature extraction \cite{xu2021automatic,wang2020u,bala2020dense}. 
Although CNN-based methods exhibit satisfactory performance in segmentation tasks, they still struggle to capture long-range dependencies and lack the ability to accurately segment targets of varying sizes.
Later, transformer-based methods were introduced due to their potential to capture global contextual relationships \cite{cao2023swin,lin2022ds,valanarasu2021medical,yuan2023effective,tragakis2023fully}. 
However, the presence of complex backgrounds and small structures in the microscopic images constrains the performance of transformer-based methods. 
Moreover, the limited data in the medical domain degrade the performance of the transformer-based methods due to their data-hungry properties. 

Recently, Mamba-based models \cite{UMamba,Segmamba,SwinUMamba} have gained attention with their ability to dynamically focus on relevant information based on the image characteristics. Numerous studies have been conducted based on the improvement of long-range modeling capabilities in CNNs when paired with Mamba blocks \cite{UMamba,lkmunet}. However, such approaches are computationally expensive and still struggle to perform better in the presence of overlapped object structures having a wide range of shape and size variations. 
Therefore, we propose a novel CNN and Mamba architecture, named InceptionMamba, which offers a simple yet effective framework for medical image segmentation tasks. As shown in Fig. \ref{fig:chart}, this framework enhances segmentation performance while significantly reducing computational complexity.
Our contributions are as follows:

\begin{itemize}
    \item We propose an efficient module, named Inception Mamba module (IMM), to capture multi-contextual representations by utilizing convolutions and state space models.
    \item We introduce a bottleneck block that effectively combines the multi-scale information from the backbone network while highlighting the fine details to improve segmentation performance.
    \item Instead of using computationally expensive dense connections, we utilize a simple decoder that reduces the computational complexity without compromising on segmentation performance.
    \item We perform extensive experimentation on four challenging medical segmentation datasets, showing that our method achieves state-of-the-art performance while reducing the computational cost by about 5 times compared to the previous state-of-the-art method, GA2-Net \cite{fiaz2024guided}. 
\end{itemize}

\section{Method}
\label{sec:method}
\vspace{-0.25cm}
This section demonstrates the proposed approach in detail. First, we briefly explain the baseline method and its limitations.  
Then we introduce the overall architecture of the proposed approach followed by an elaboration on our Inception Mamba module (IMM) that efficiently captures the multi-contextual features and provides rich semantic information necessary for the segmentation task. Finally, we discuss the proposed bottleneck block that effectively combines the multi-scale information from backbone stages while preserving the boundary information.
\vspace{-0.3cm}
\subsection{Baseline Approach}
\label{ssec:baeline_approach}
We adopt a U-Net~\cite{ronneberger2015u} based architecture as our baseline model.
Specifically, the baseline model consists of a ResNet \cite{he2016deep}  backbone followed by a decoder that combines the high-level semantic information with multi-scale features from the backbone in a U-shaped fashion.
Despite utilizing the multi-scale information from the backbone, the baseline method struggles to accurately segment the complex object structures having varying shapes and sizes in microscopic medical images. Additionally, the presence of background artifacts and clutter in the images further complicate the segmentation task of the various types of medical objects such as tissues and cells. 

To this end, we propose an Inception Mamba module (IMM) that efficaciously captures the multi-contextual features to segment complex-shaped medical objects of different sizes. Furthermore, our proposed bottleneck block effectively combines the multi-scale information from the backbone network while highlighting the boundary information for better segmentation of targets in various backgrounds. Next, we discuss the overall architecture of the proposed model in detail.
\vspace{-0.25cm}
\begin{figure*}[t!]
\centering
 \includegraphics[width=0.9\linewidth]{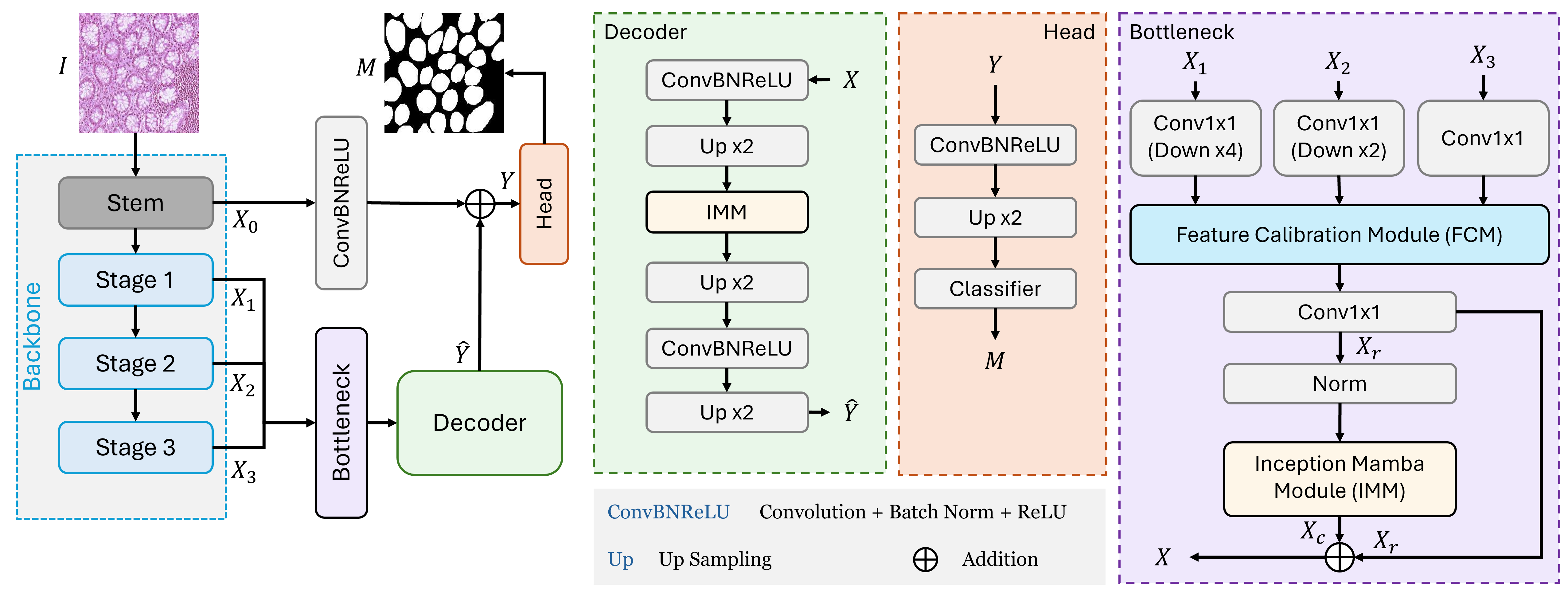} 
\vspace{-0.25cm}
 \caption{Illustration of the overall architecture of our proposed framework. \textbf{(left)} We feed the image to the backbone and extract features $X_i$ where $i \in {0,1,2,3}$. Multi-stage features from the backbone are effectively combined and enhanced by the bottleneck block. Afterwards, the decoder upsamples and further refines multi-contextual features. Finally, the output of the decoder is combined with features of the stem, and the segmentation head is utilized to obtain mask $M$. \textbf{(middle)} We also illustrate the structure of the decoder (light green box) and segmentation head (light orange box). \textbf{(right)} The bottleneck block, which includes the Feature Calibration Module, is shown in light purple.} 
 \vspace{-0.5cm}
    \label{fig:overall_arch}
\end{figure*}

\subsection{Overall Architecture}
\label{ssec:overall_arch}
We illustrate the overall architecture of the proposed framework in Fig. \ref{fig:overall_arch}. As depicted, the model takes an image $I \in \mathbb{R}^{C \times H \times W}$ as input and feeds it to the ResNet backbone to extract multi-scale features $X_i$ where $i = {0,1,2,3}$. 
Here, $i = 0$ refers to the features of the stem layer. 
To reduce the computational complexity of the model, we utilize the features of the first three stages of the ResNet backbone. We then pass the multi-scale features to the bottleneck block for enrichment. 
The bottleneck block takes the features of three stages and passes them to a feature calibration module (FCM) that is responsible for highlighting the boundary information by utilizing the low and high-frequency information in a parallel manner. 
Afterward, IMM is utilized to capture the multi-contextual features from the refined feature maps. 
The proposed bottleneck block provides the rich multi-contextual information necessary for better segmentation performance. 

The output of the bottleneck block is input to a decoder that uses several convolution and upsampling layers in a cascade manner. We carefully utilize the IMM in the decoder to further enhance the multi-contextual information as shown in Fig. \ref{fig:overall_arch} (light green box). Finally, we add the low-level semantics of the stem layer of the backbone network with the semantically rich feature maps of the decoder and feed it to a segmentation head module (as shown in Fig. \ref{fig:overall_arch} (light orange box)) which provides the output prediction mask $M$.
\begin{wrapfigure}{r}{0.45\textwidth}
  \centering
  \vspace{0.4cm}
 \includegraphics[width=1\linewidth]
 {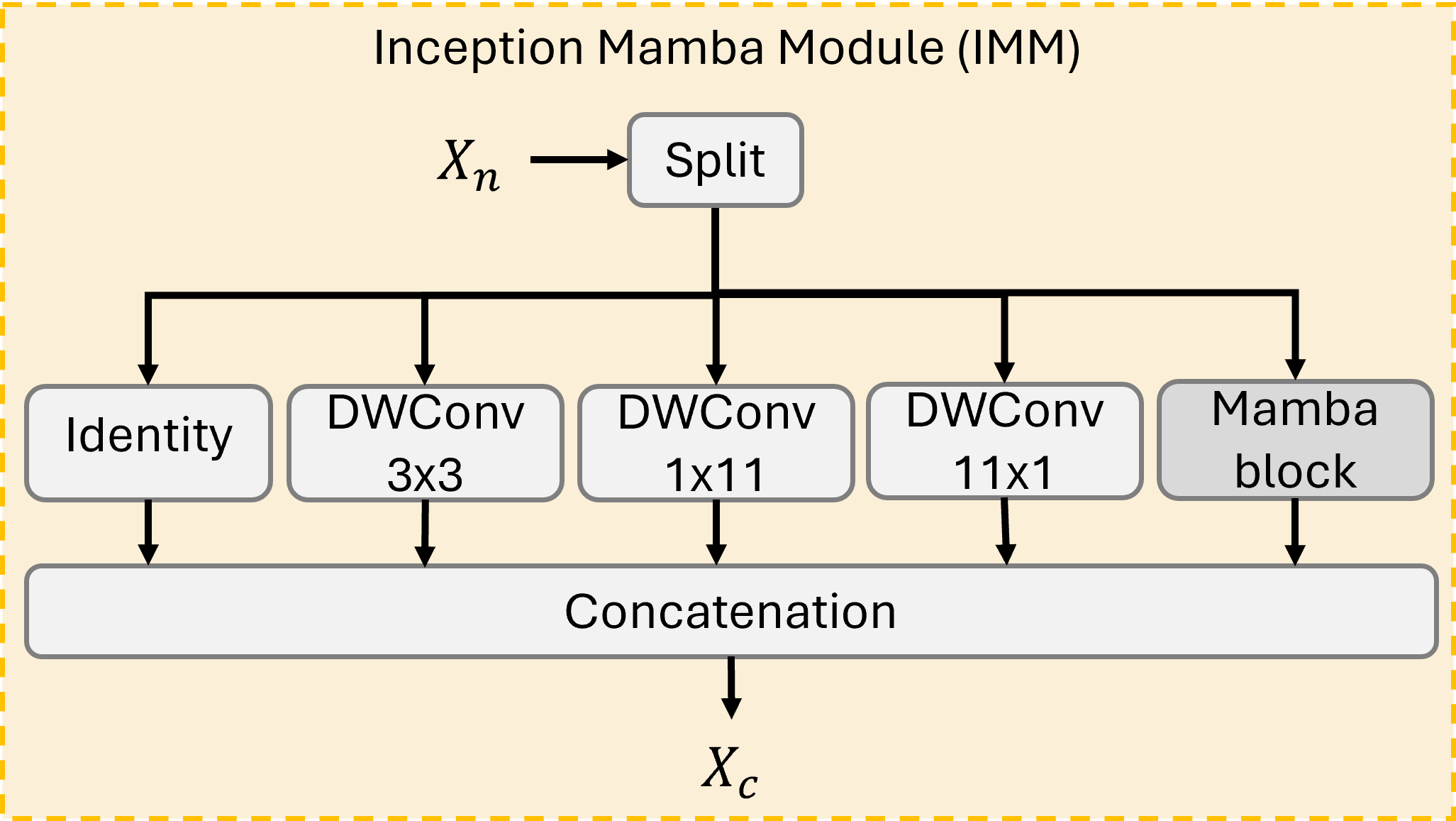} 
 \caption{Inception Mamba module (IMM) takes input features $X_n$, splits them channel-wise, and utilize various convolution kernels and state space module to extract multi-contextual features $X_c$ for better segmentation performance as depicted above.}  
    \label{fig:inception_module}
    \vspace{-0.8cm}
    \end{wrapfigure}
\vspace{-0.5cm}
\subsection{Inception Mamba Module}

The Inception Mamba module (IMM) is illustrated in Fig. \ref{fig:inception_module}. As mentioned earlier,
medical microscopic images contain complex object structures having varying shapes and sizes. Therefore, an explicit module is required to effectively detect medical objects in microscopic images. 
To this end, we introduce a simple and effective module, the Inception Mamba module (IMM), that simultaneously captures different contextual representations while aiming to minimize the computational complexity. The input feature $X_n$ is split channel-wise into multiple subsets to process them through different branches. The subsets  $X_{n0}$,  $X_{n1}$,  $X_{n2}$,  $X_{n3}$,  $X_{n4}$ are passed for identity and depth-wise convolution operations while $X_{n4}$  is input to the mamba block. During the depth-wise convolutions, square and rectangular convolutional kernels are utilized to obtain feature representations having smaller and larger receptive fields. We choose depth-wise convolutions to build a computationally efficient module. In parallel, the state-space Mamba block is employed to capture larger contextual representations while reducing the computational cost. 
In contrast to the Inception block \cite{szegedy2015going,szegedy2016rethinking} that uses the max-pooling operation to add another context branch, we utilize the identity operation. We empirically observe that the identity branch when utilized in IMM provides better segmentation performance. 
Finally, we concatenate the different representations in the channel dimension to obtain the rich multi-contextual feature maps ($X_c$).

\noindent\textbf{Mamba Block: }
Mamba \cite{gu2023mamba} effectively encodes global representations while benefiting from low computational costs, striking a balanced trade-off between performance and efficiency.
In addition, the selective scan mechanism of Mamba has the potential to mine the core semantics for long sequences which reduces the semantic redundancy limitation.
Motivated by the selective scan mechanism and high efficiency
of the Mamba block, we integrate it into our inception Mamba module to capture the various contextual representations. This fusion of the Mamba module within our IMM provides a mutually beneficial outcome by capturing the global representations while maintaining high efficiency.
\vspace{-0.25cm}

\subsection{Bottleneck Block}
\begin{wrapfigure}{r}{0.45\textwidth}
\vspace{-0.6cm}
  \centering
\includegraphics[width=0.8\linewidth]{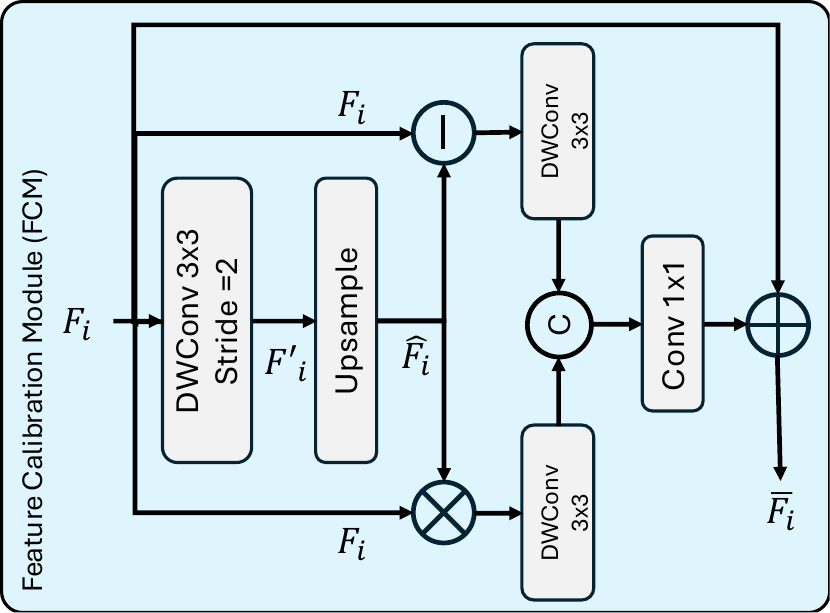} 
 \caption{Illustration of the feature calibration module. Feature maps $F_i$ are first down-sampled using depth-wise convolution followed by an upsampling operation. Then, we multiply features $F_i$ with $\hat{F_i}$ to highlight blob regions. Similarly, subtraction of $F_i$ and $\hat{F_i}$ is performed to focus on fine details. Finally, highlighted features are combined and projected using convolution operation to get rich semantic features.} 
 \vspace{-0.3cm}
    \label{fig:feat_calib}
    \end{wrapfigure}
The critical component of the proposed framework is the bottleneck block as shown in Fig. \ref{fig:overall_arch} (right). The bottleneck block takes the features maps, $X_i$ where $i \in {1,2,3}$, of three stages from the backbone network and enhances them by employing the feature calibration module (FCM). 
The FCM is responsible for highlighting the fine details that are necessary to separate the overlapped cells and tissues as well as enhancing the feature contrast. 
The feature representations of each stage are separately enhanced by the FCM blocks. For the first stage, i.e. $i = 1$, feature maps are first downsampled by a factor of four before inputting them into the FCM block. 
Similarly, we downsample the feature representations of the second stage by a factor of two and then feed them to the FCM block. 
After feature enhancement, we concatenate the feature maps of three stages and utilize the $1 \times 1$ convolution operation to intermix the different responses and obtain rich multi-scale representations $X_r$. 
Later, we utilize the IMM to further boost the contextual representations and obtain features $X_c$. Finally, we add the features $X_r$ and $X_c$ to obtain semantically rich features. 
Both FCM and IMM complement each other to provide better representations for medical segmentation tasks. 
Our approach effectively combines the low-level semantics of the first and second stages with the high-level semantics of the third stage of the backbone network to achieve better segmentation performance.


\noindent\textbf{Feature Calibration Module: } 
As shown in Fig. \ref{fig:feat_calib}, the feature calibration module takes input features $F_i$ and utilizes a $3 \times 3$ depth-wise convolution having a stride value of 2 to reduce the spatial resolution of the features producing ${F'}_i$. We then use interpolation operation to upsample the spatial resolution of the features to obtain feature maps $\hat{F}_i$. The objective of the convolutional down sampling followed by upsampling is to produce smoothness in the feature maps. Subsequently, we subtract the features $\hat{F}_i$ from the $F_i$ to highlight the fine details. Simultaneously, we multiply the features $\hat{F}_i$ and $F_i$ to focus on the blob regions. Afterwards, we utilize the convolution operation on the subtracted and multiplied features and concatenate them. Finally, we utilize the $1 \times 1$ convolution to project the feature representations and add the input $F_i$ to obtain enhanced feature maps $\Bar{F}_i$. The feature calibration module enhances boundary information
addressing the complex nature of medical image segmentation.
\vspace{-0.3cm}
\subsection{Decoder}
To reduce the computational cost of the framework, we introduce a simple and effective decoder. We empirically verify that the dense skip connections in the baseline increase the computational cost while negligibly improving the segmentation performance. Additionally, we observe that the high-level semantics of the fourth stage of the backbone are not profitable for microscopic medical images. We therefore utilize a decoder that does not require skip connections while still achieving better segmentation performance. The proposed decoder (as shown in Fig. \ref{fig:overall_arch} light green box) takes the output of the bottleneck block and utilizes the convolution and upsampling layers to obtain features $\hat{Y}$. Then we enhance the feature maps by using the multi-scale contextual module IMM followed by an upsampling operation. Finally, we use the convolution and upsampling operation again to get the feature maps having half the spatial resolution of the input image which is then passed to the head for segmentation prediction.
\vspace{-0.4cm}
\section{Experimental Study}

\subsection{Datasets \& Experimental details}
\begin{wraptable}{r}{5.7cm}
\vspace{-2.5cm}
\centering
\scalebox{0.56}{
\begin{tabular}{l|c|c|c|c}
\hline
\multicolumn{1}{c}{\multirow{2}{*}{Method}} & \multicolumn{1}{|c}{\multirow{2}{*}{Params (M)}} & \multicolumn{1}{|c}{\multirow{2}{*}{GFLOPs}} & \multicolumn{2}{|c}{SegPC21}   \\ \cline{4-5} 
\multicolumn{1}{c}{}  & \multicolumn{1}{|c}{}  & \multicolumn{1}{|c}{}  & \multicolumn{1}{|l|}{Dice (\%)} & \multicolumn{1}{l}{IoU (\%)}  \\ \cline{4-5} 
\hline
U-Net~\cite{ronneberger2015u}    & {14.8} & 50.3 & 88.08 & 88.2 \\
UNet++~\cite{zhou2018unet++}  & 74.5 & 94.6 & 91.02 & 90.92 \\
AttUNet~\cite{oktay2018attention}  & 34.9 & 101.9 & 91.58 & 91.44 \\
MultiResUNet~\cite{ibtehaz2020multiresunet} & 57.2 & 78.4  & 86.49 & 86.76  \\

TransUNet~\cite{chen2021transunet} & 105.0 & 56.7 & 82.33 & 83.38  \\
MissFormer\cite{huang2021missformer} & 42.46  & 9.86 & 80.82 & 82.09\\
UCTransNet~\cite{wang2022uctransnet}& 65.6 & 63.2 & 91.74 & 91.59  \\
SA2-Net \cite{fiaz2023sa2} & 19.3 & 35.36 & 92.41 & 92.23 \\
DwinFormer \cite{kareem2024medical}  & 198.0 & 113.13 & {91.10}  & {90.99}  \\
UDTransNet \cite{wang2024narrowing}  & 33.8  & 63.2 &  89.91  & - \\
U-Mamba \cite{UMamba}& \textcolor{blue}{12.36} & 51.5 & 91.79 & 91.83\\
LKMUNet \cite{lkmunet}& 123.8 & 251.5 & 92.20 & 92.02 \\

GA2-Net \cite{fiaz2024guided}& 17.36 & 32.61 & 92.49 & 92.32 \\
\hline
\textbf{(Ours)} & \textcolor{red}{11.92} & \textcolor{blue}{6.72} & \textcolor{blue}{92.56}  & \textcolor{blue}{92.37} \\
\textbf{(Ours$^*$ )} & 27.27 & \textcolor{red}{4.86} & \textcolor{red}{92.84}  & \textcolor{red}{92.63} \\
\hline
\end{tabular}}
\caption{Comparison of our method on SegPC21 dataset with state-of-the-art methods. The best two results are in \textcolor{red}{red} and \textcolor{blue}{blue}, respectively. $\star$  means the model backbone is PVT-V2-B2 (\protect\cite{wang2022pvt}).}
\vspace{-0.5cm}
\label{tab:segpc}
\end{wraptable}

\textbf{Multiple Myeloma Segmentation: }
Multiple myeloma (MM) is known as a plasma cancerous cell in the blood. 
The SegPC21 \cite{gupta2021segpc} dataset is a collection of 775 bone marrow aspirate slide images of MM patients. We split the dataset into train, val, and test sets comprising 290, 200, and, 277 samples, respectively. Similar to \cite{fiaz2023sa2,azad2022medical}, we perform the cytoplasm segmentation task after cropping the nucleus samples.
\noindent\textbf{Gland Segmentation: }
In colon histology challenge, Gland segmentation dataset (GlaS) ~\cite{sirinukunwattana2017gland} was introduced to segment the Gland. It comprises 85 training samples and 80 test cases that are captured from 16 H\&E stained histological sections of colorectal adenocarcinoma. 
 \noindent\textbf{Skin Lesion Segmentation: }
Apart from microscopic images, we also perform skin lesion segmentation over dermoscopic images and use two commonly used datasets i.e., ISIC2018 \cite{codella2019skin} and ISIC2017 \cite{codella2018skin}. The ISIC 2017 dataset comprises 2000 training images and has 150 and 600 samples used for validation and test purposes, respectively. On the other hand,  for ISIC2018, we follow \cite{fiaz2023sa2,azad2022medical} to obtain training, validation, and test splits of 1815, 259, and 520 images, respectively.

In this study, we perform all the experiments using 32G Tesla V100 GPU based on PyTorch 2.1.1+cu118. We utilize ResNet50 \cite{he2016deep},  pre-trained on ImageNet \cite{deng2009imagenet}, as our backbone network to extract the features from the stem layer and the first three-layer features. We set the input resolution to $224 \times 224$.

\begin{wraptable}{r}{5.2cm}
\vspace{-0.7cm}
\centering
\scalebox{0.7}{
\begin{tabular}{l|c|c}
\hline
\multicolumn{1}{c}{\multirow{2}{*}{Method}} & \multicolumn{2}{|c}{GlaS}   \\ \cline{2-3} 
\multicolumn{1}{c}{}  & \multicolumn{1}{|l|}{Dice (\%)} & \multicolumn{1}{l}{IoU (\%)}\\ \cline{2-3} 
\hline
U-Net~\cite{ronneberger2015u}      & 85.45 $\pm$ 1.3 & 74.78 $\pm$ 1.7  \\
UNet++~\cite{zhou2018unet++}      & 87.56 $\pm$ 1.2  & 79.13  $\pm$ 1.7   \\
AttUNet~\cite{oktay2018attention}     & 88.80 $\pm$ 1.1   & 80.69 $\pm$ 1.7  \\
MultiResUNet~\cite{ibtehaz2020multiresunet}  & 88.73 $\pm$ 1.2  & 80.89 $\pm$ 1.7   \\ 
TransUNet~\cite{chen2021transunet}   & 88.40 $\pm$ 0.7  & 80.40$\pm$ 1.0   \\
MedT~\cite{valanarasu2021medical}        & 85.93 $\pm$ 2.9  & 75.47 $\pm$ 3.5  \\
Swin-Unet~\cite{cao2023swin}   & 89.58 $\pm$ 0.6  & 82.07 $\pm$ 0.7 \\
UCTransNet~\cite{wang2022uctransnet}  & 90.18 $\pm$ 0.7  & 82.96 $\pm$ 1.1  \\
 SA2-Net  \cite{fiaz2023sa2} & \textcolor{blue}{91.38 $\pm$ 0.4}  & \textcolor{blue}{84.90 $\pm$ 0.6}  \\
UDTransNet \cite{wang2024narrowing} & 91.03  $\pm$  0.6  & - \\ \hline
  \textbf{Ours}   & \textcolor{red}{91.88 $\pm$ 0.3}  & \textcolor{red}{85.65 $\pm$ 0.4}  \\
\hline
\end{tabular}}
\caption{Comparison of our method on the GlaS dataset with state-of-the-art methods. The best two results are in \textcolor{red}{red} and \textcolor{blue}{blue}, respectively.}
\vspace{-0.5cm}
\label{tab:glas}
\end{wraptable}
During the training, we apply rotation and random flipping augmentation techniques.
We train the model using the combined cross-entropy and DICE loss functions. For SegPC21, ISIC2018, and ISIC2017 datasets, we follow \cite{fiaz2023sa2,azad2022medical} and set batch size of 16, learning rate 0.0001, and use the Adam optimizer to train the model for 100 epochs. For the GlaS dataset, we follow  UCTransNet \cite{wang2022uctransnet} and set the batch size to 4, the initial learning rate to 0.001, and employ the Adam optimizer to train our model. 

To ensure the results are robust for a smaller dataset, similar to \cite{fiaz2023sa2,wang2022uctransnet}, we perform three times 5-fold cross-validation. 
During inference, we employ an ensemble technique to obtain the final prediction masks by taking the mean for all five models. We evaluate our method using the Dice and IoU metrics.

\begin{wraptable}{r}{5.9cm}
\vspace{-0.99cm}
\scalebox{0.59}{
\centering
\begin{tabular}{l|c|c|c|c}
\hline
\multicolumn{1}{c}{\multirow{2}{*}{Method}} & \multicolumn{2}{|c|}{ISIC2017} & \multicolumn{2}{c}{ISIC2018}   \\ \cline{2-5} 
\multicolumn{1}{c}{}  & \multicolumn{1}{|l|}{Dice (\%)} & \multicolumn{1}{l|}{IoU (\%)} & \multicolumn{1}{l|}{Dice (\%)} & \multicolumn{1}{l}{IoU (\%)} \\ \cline{2-5} 
\hline
U-Net~\cite{ronneberger2015u}      & 81.59 & 79.32 & 86.71 & 84.91 \\
UNet++~\cite{zhou2018unet++}    & 82.32 & 80.13 & 88.22 & 86.51 \\
AttUNet~\cite{oktay2018attention}    & 86.45 &85.97 & 88.20 & 86.49 \\
MultiResUNet~\cite{ibtehaz2020multiresunet}    & 86.83& 85.91 & 86.94 & 85.37 \\

TransUNet~\cite{chen2021transunet}  & 86.11 & 84.98 & 84.99 & 83.65 \\
MissFormer\cite{huang2021missformer}  & 87.82 & 85.79& 86.57 & 84.84 \\
UCTransNet~\cite{wang2022uctransnet} & 88.81& 87.22 & {88.98} & {87.29} \\
SA2-Net \cite{fiaz2023sa2}   & 89.59 & 88.14  & {88.88}  & {87.21} \\
UDTransNet \cite{wang2024narrowing}  & 90.11 & 88.49  &  \textcolor{blue}{89.91}  & 88.02 \\
U-Mamba \cite{UMamba}   & {89.60}  & {88.45}  & {89.37}  & \textcolor{blue}{88.12} \\
LKMUNet \cite{lkmunet}   & \textcolor{blue}{90.25}  & \textcolor{blue}{88.89}  & {89.33}  & {87.60} \\
GA2-Net \cite{fiaz2024guided}   & 89.99  & {88.58}  & {89.29}  & {87.69} \\
\hline
\textbf{Ours} & \textcolor{red}{91.16}  & \textcolor{red}{89.82}   & \textcolor{red}{90.56}  & \textcolor{red}{88.98} \\
\hline
\end{tabular}}
\caption{Comparison with state-of-the-art methods on ISIC2017 and ISIC2018 datasets using the Dice and IoU metrics.  The best two results are in \textcolor{red}{red} and \textcolor{blue}{blue}, respectively.}
\vspace{-0.3cm}
\label{tab:isic2017-isic2018}
\end{wraptable}
\vspace{-0.5cm}
\subsection{Quantitative Comparison}
We compare our method with existing CNN-based, transformer-based, and Mamba-based methods on the SegPC21 dataset in Table \ref{tab:segpc}. We report Dice score, IoU, number of parameters (in millions), and FLOPs (GFLOPs). From Table \ref{tab:segpc}, compared to CNN methods, we notice that AttUNet \cite{oktay2018attention} attains the best Dice and IoU scores of 91.58\% and 91.44\%, respectively. However, among hybrid approaches, GA2Net \cite{fiaz2024guided} achieves 92.49\% Dice and 92.32\% IoU scores. In contrast, our approach achieves state-of-the-art Dice and IoU scores of 92.56\% and 92.37\%, respectively. In addition, we notice that our approach obtains a significant amount of reduced parameters and GFLOPs. Compared to  GA2Net, our method has around 1.45 times fewer parameters and 4.85 times fewer GFLOPs. Moreover, we also demonstrate that while employing the PVT-V2-B2 \cite{wang2022pvt} backbone, our method further improves the Dice and IoU scores to 92.84\% and 92.63\%, respectively. We notice that this further reduces the computational cost to 4.86 GFLOPs. 
Furthermore, our method achieves an impressive HD95 score of 1.92 ± 0.32 compared to the HD95 score of 3.87 ± 3.56 for the top-performing GA2Net model which indicates better boundary segmentation capability.

In Table~\ref{tab:glas}, we compare our method on the GlaS dataset. 
Among the compared methods, we observe that SA2Net \cite{fiaz2023sa2} and UDTransNet \cite{wang2024narrowing} attain around 91.00\% Dice score. In contrast, we achieve 91.88\% Dice and 85.65\% IoU score which demonstrates the state-of-the-art performance of our approach. We also compare our method on ISIC2017 and  ISIC2018 datasets in Table~\ref{tab:isic2017-isic2018}. Among CNN-based methods,  MultiResUNet \cite{ibtehaz2020multiresunet} ranks first with a Dice score of 86.94 \% and an IoU score of 85.37 on ISIC2018 dataset. However, the recently introduced UDTransNet \cite{wang2024narrowing}, GA2Net \cite{fiaz2024guided}, U-Mamba \cite{UMamba}, and LKMUNet \cite{lkmunet} achieve a Dice score of more than 89.00\%. Our approach demonstrates better performance with a Dice score of 90.56\% and an IoU score of 88.98\% on ISIC2018 dataset.  
We have computed the HD95 score for our proposed method. On the ISIC2018 dataset, our method achieves an HD95 score of 4.15 ± 0.20 (where lower is better), significantly outperforming the state-of-the-art UDTransNet, which has an HD95 score of 10.85 ± 0.26.
 Similarly, our method shows consistent improvement on ISIC2017. 

\subsection{Qualitative Comparison}
\begin{wrapfigure}{r}{0.65\textwidth}
\centering
 \vspace{-0.5cm}
 \includegraphics[width=0.9\linewidth]{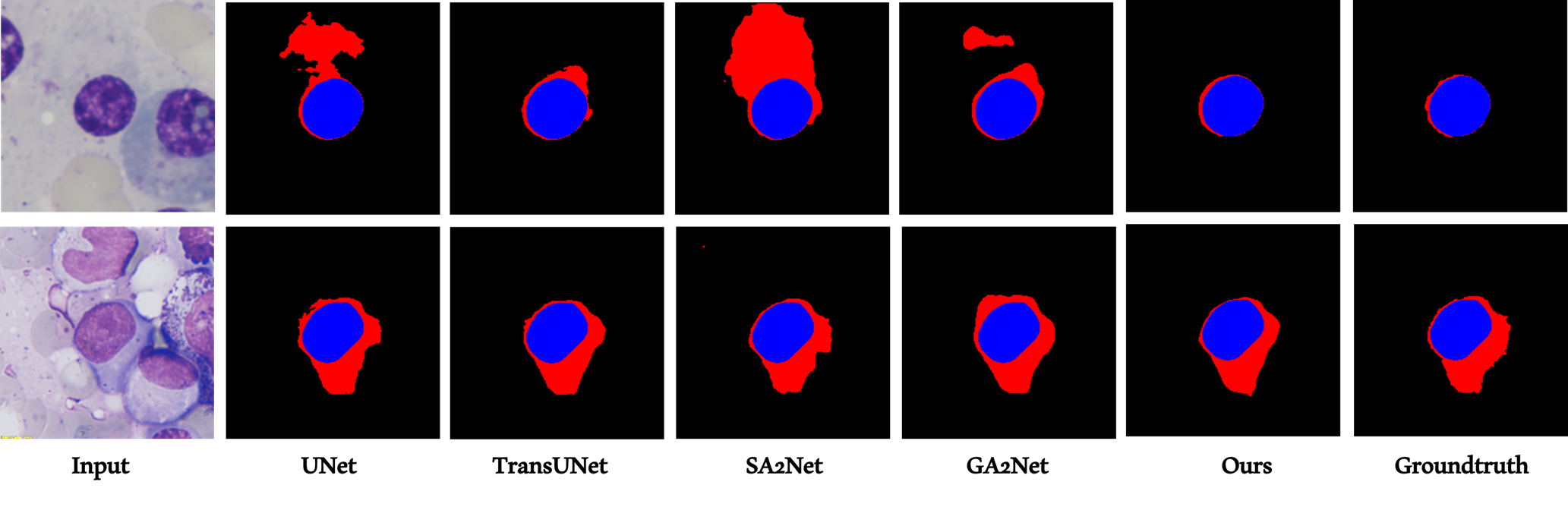} 
  \vspace{-0.2cm}
 \caption{Qualitative comparison of the proposed method on the {SegPC21} dataset samples. For a fair comparison, we follow \protect\cite{fiaz2023sa2,azad2022medical} and perform the cytoplasm segmentation (\textcolor{red}{red mask}) for a given input nucleus mask (\textcolor{blue}{blue}). Our method provides improved segmentation performance by accurately detecting the cytoplasm region with clear boundaries, compared to existing methods.  } 
 \vspace{-0.3cm}
    \label{fig:segpc} 
\end{wrapfigure}



\begin{wrapfigure}{r}{0.7\textwidth}
\centering
\vspace{-0.5cm}
 \includegraphics[width=0.9\linewidth]{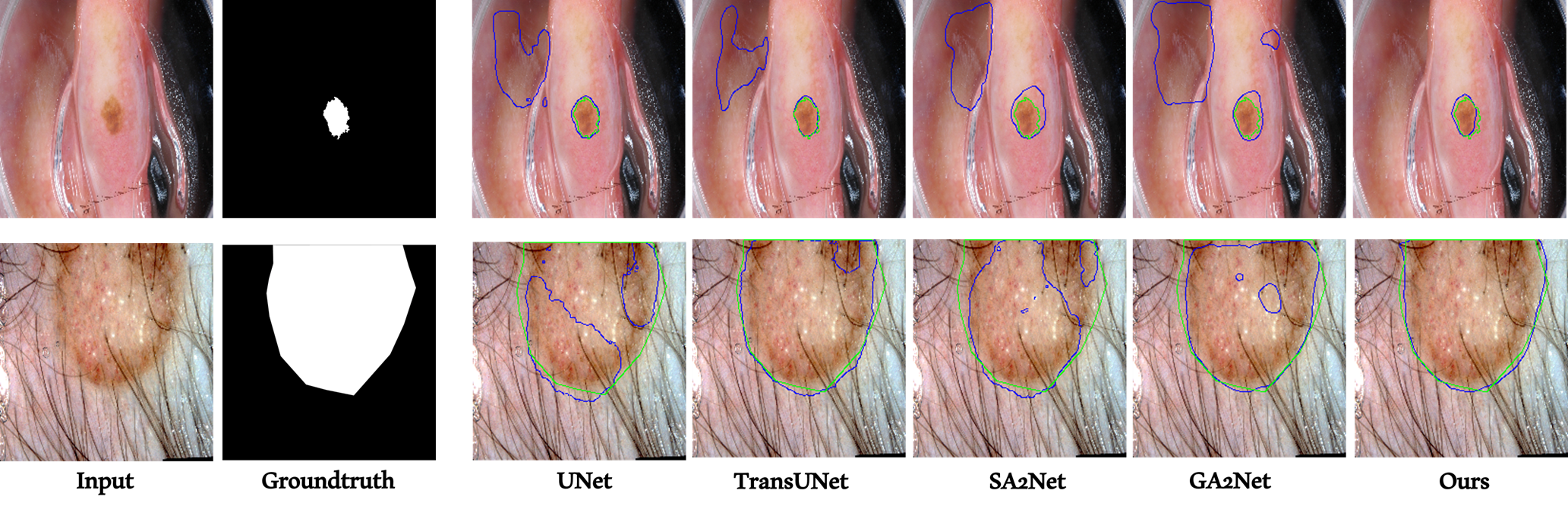} 
  \vspace{-0.5cm}
\caption{Qualitative comparisons of different methods on the ISIC2018 skin lesion samples. Ground truth boundaries are shown in (\textcolor{green}{green}) and predicted boundaries are shown in (\textcolor{blue}{blue}). We observe that our method performs better in cluttered backgrounds and correctly detects the affected region in the blue, compared to existing methods.}
\vspace{-0.3cm}
 \label{fig:ISIC2018}
\end{wrapfigure}

To further demonstrate the effectiveness of our method, we perform a qualitative comparison in Figures \ref{fig:segpc} and \ref{fig:ISIC2018}.
We compare our method with UNet \cite{ronneberger2015u},  TransUNet \cite{chen2021transunet}, SA2Net \cite{fiaz2023sa2} and GA2Net \cite{fiaz2024guided}. 
In Fig. \ref{fig:segpc}, we compare the methods on SegPC21 dataset samples and observe that our method exhibits better capabilities to learn the multi-scale features and preserve the boundary regions of the cytoplasm in the cluttered and unclear backgrounds.
In addition, we notice that our method has better learning capabilities to capture the underlying complex skin lesion tissues in Fig. \ref{fig:ISIC2018}. This indicates that our method has the potential to grasp the intricate details of the objects, can preserve the contours, and shows robustness against the noisy items in the image. 
This can be attributed to the expedient combination of our proposed feature calibration module and Inception Mamba module within the bottleneck to refine the features and capture the contextual information from multi-stage features.




\subsection{Ablation Study}

In Table ~\ref{ablation}, we present an ablation study on SegPC21 dataset to validate the effectiveness of our contributions. We adopt modified  UNet \cite{ronneberger2015u} without skip connections as the baseline, pre-trained on ImageNet \cite{deng2009imagenet}.
The baseline utilizes the stem features and the first three stages feature as multi-stage features, which are fused and passed to a decoder with upsampling and convolution layers. \\
\begin{wraptable}{r}{9.5cm}
\centering
\scalebox{0.63}{
\begin{tabular}{
c 
c 
c 
c 
c }
\hline
Exp. No. & Methods                                                        & Dice                         & Params(M)                    & GFLOPs                      \\ \hline
1 & Baseline                                                       & 89.05                        &  {\color[HTML]{333333} 11.42} & {\color[HTML]{333333} 5.27} \\ 
2 & Baseline + FCM                                                 & {\color[HTML]{333333} 90.4}  & {\color[HTML]{333333} 11.53} & {\color[HTML]{333333} 5.29} \\ 
3 & Baseline +IMM (with only Inception DWC)                                                 & {\color[HTML]{333333} 89.6}  & {\color[HTML]{333333} 11.42} & {\color[HTML]{333333} 5.27} \\ 
4 & Baseline +IMM (with only Mamba)                                                 & {\color[HTML]{333333} 90.2}  & {\color[HTML]{333333} 11.42} & {\color[HTML]{333333} 5.70} \\ 
5 & Baseline +IMM (with Inception DWC and Mamba)                                                 & {\color[HTML]{333333} 90.9}  & {\color[HTML]{333333} 11.43} & {\color[HTML]{333333} 5.28} \\ 
6 & Baseline+ Inception Module \cite{szegedy2015going}                            & {\color[HTML]{333333} 89.4}  & {\color[HTML]{333333} 11.43} & {\color[HTML]{333333} 5.27} \\ 
7 & Baseline+ Inception Module \cite{szegedy2016rethinking}                             & {\color[HTML]{333333} 89.3}  & {\color[HTML]{333333} 11.43} & {\color[HTML]{333333} 5.27} \\ 
8 & Baseline+ Inception Module \cite{szegedy2016rethinking}                           & {\color[HTML]{333333} 89.5}  & {\color[HTML]{333333} 11.44} & {\color[HTML]{333333} 5.29} \\ 
9 & Baseline+ Self-Attention (SA)                                  & {\color[HTML]{333333} 89.9}  & {\color[HTML]{333333} 12.07} & {\color[HTML]{333333} 6.7}  \\ 
10 & Baseline+IMM (SA replacing Mamba)                        & {\color[HTML]{333333} 90.3}  & {\color[HTML]{000000} 12.46} & {\color[HTML]{333333} 8.3}  \\ 
11 & Baseline + FCM + IMM                                           & {\color[HTML]{333333} 91.5}  & {\color[HTML]{333333} 11.54} & {\color[HTML]{333333} 5.30} \\ 
12 & {\color[HTML]{333333} Baseline + FCM + IMM + Decoder with IMM} & {\color[HTML]{333333} 92.05} & {\color[HTML]{333333} 11.67} & {\color[HTML]{333333} 6.09} \\ 
13 & Ours                                                           & {\color[HTML]{333333} 92.56} & {\color[HTML]{333333} 11.92} & {\color[HTML]{333333} 6.72} \\ \hline
\end{tabular}
}
\caption{ Ablation Study comparing the Dice score and computational requirements of our method over SegPC21 dataset. It is notable that  FCM and IMM modules in the bottleneck and decoder positively impact the model performance with lesser computational cost. DWC refers to Depth-wise Convolutions.}
\vspace{-0.3cm}
\label{ablation}
\end{wraptable}

We experiment combinations of different modules with baseline including FCM (Exp 2),  IMM with only Inception depth-wise convolutions (Exp 3), IMM with only Mamba block (Exp 4), the proposed IMM (Exp 5), variants of Inception Module (Exp 6 - 8), Self-Attention (Exp 9) and IMM utilizing Self-Attention instead of Mamba block (Exp 10). It is evident that baseline combined with FCM  (Exp 2)  and with IMM (Exp 5) outperform other combinations with comparable parameters and GFLOPs. Although Self-Attention and Inception Modules are capable of encoding global details, they fail to explicitly encode complex semantic details for identification of targets with varying shapes and blurred boundaries. Hence, we integrate the IMM and FCM modules in to baseline which is set as our design choice. Moreover, directly using the convolutions and upsampling in a decoder might lose complementary information.
We modified the decoder and introduced our IMM module within the decoder which resulted in a significant gain in performance (Exp 12). Finally, we utilize a skip connection from stem features using a convolution layer which serves a complimentary function, which further improves the performance (Exp 13).

\begin{figure*}[t!]
\centering
 \includegraphics[width=0.8\linewidth]{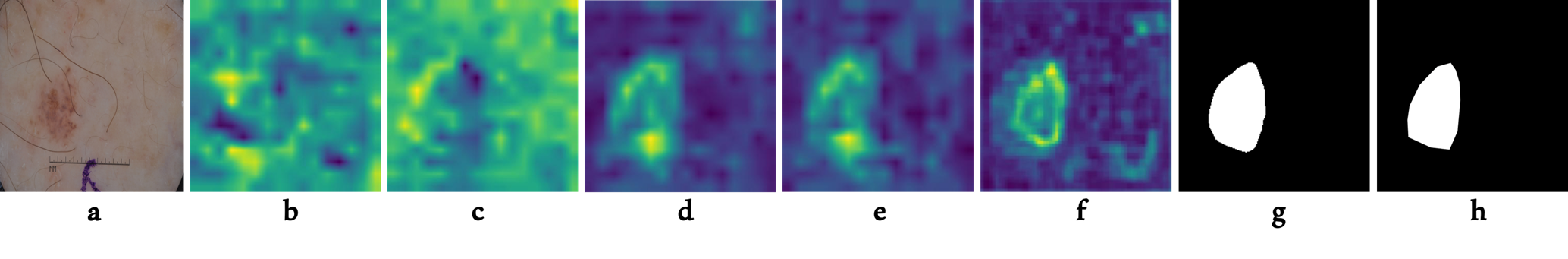} 
 \vspace{-0.3cm}
\caption{Illustration of the impact of our contributions. (a) is the input to the model, (b) and (c) presents the output
feature maps of stage 2 and stage 3 features. The (d) and (e) are output features by FCM and IMM, respectively. The (f) is the feature map of our decoder. Finally (g) and (h) are our model prediction and ground truth, respectively. These feature visualizations show that our model can capture intricate details and preserve the precise contours for the unclear boundaries of the tissues.}
 \vspace{-0.5cm}
 \label{fig:contribution_impact}
\end{figure*}

\begin{wraptable}{r}{6cm}
\vspace{-0.4cm}
\centering
\scalebox{0.83}{
\begin{tabular}{l c}
\hline
Method  & Dice\\ \hline
First place of IMM in decoder     & 92.21    \\
Second place of IMM in decoder    & \textbf{92.56}    \\
Third place of IMM in decoder     & 92.32    \\
\hline
\end{tabular}}
\caption{We periodically replace convolution layers in decoder with IMM and show the results over the SegPC21 dataset in terms of Dice score. We notice that placing IMM in the middle of decoder results in the optimal solution.}
\vspace{-0.5cm}
\label{tab:abs_decoder}
\end{wraptable}
In addition, we carefully analyze the impact of the IMM design by evaluating the individual contributions of the Inception DWC and Mamba blocks within the IMM architecture (Exp 3, Exp 4). It can be noticed that integration of IMM (with only Inception DWC) into the baseline  improves the performance while maintaining the computational cost, whereas the integration of IMM (with only Mamba block) increases both the performance and computational cost due to the large number of channels for the Mamba block. When both the Inception DWC and Mamba modules of the IMM block are included (Exp 11), the model achieves further Dice score improvement (+1.85\%) over the baseline, while maintaining a computational cost (GFLOPs) nearly identical to the baseline. Note that splitting the feature channels between the Identity, Inception DWC, and Mamba blocks within the IMM enhances performance while limiting the computational cost compared to using Mamba blocks alone. From these experiments, it is clear that our contributions exhibit better capabilities to learn refined multi-context information for accurate segmentation tasks. Furthermore, we  perform an ablation study to find the optimal position of our IMM in the decoder. 
In Table~\ref{tab:abs_decoder}, it can be seen that IMM in the middle of the decoder provides optimal performance gain. Finally, we show the feature map visualizations of our contribution in Fig. \ref{fig:contribution_impact}, which highlights that our model can learn the complex structures of tissues in the presence of cluttered and noisy backgrounds.
 \vspace{-0.5cm}

\section{Conclusion}
In this paper, we propose a segmentation framework that exploits multi-stage features from a backbone architecture and passes them to the bottleneck which is responsible for generating enriched multi-contextual feature representations using feature calibration and Inception Mamba modules. 
The proposed Inception Mamba Module (IMM) captures the multi-contextual feature representations while remaining computationally efficient.
Furthermore, 
our simple yet effective decoder incorporating Inception Mamba module provides better segmentation results. Experimental results on two microscopic and two skin lesion segmentation datasets reveal the significance of our approach.

\bibliography{egbib}

\begin{thebibliography}{35}
\providecommand{\natexlab}[1]{#1}
\providecommand{\url}[1]{\texttt{#1}}
\expandafter\ifx\csname urlstyle\endcsname\relax
  \providecommand{\doi}[1]{doi: #1}\else
  \providecommand{\doi}{doi: \begingroup \urlstyle{rm}\Url}\fi

\bibitem[Azad et~al.(2024)Azad, Aghdam, and et~al.]{azad2022medical}
Reza Azad, Ehsan~Khodapanah Aghdam, and et~al.
\newblock Medical image segmentation review: The success of u-net.
\newblock \emph{IEEE Transactions on Pattern Analysis and Machine Intelligence}, 2024.

\bibitem[Bala and Kant(2020)]{bala2020dense}
Surayya~Ado Bala and Shri Kant.
\newblock Dense dilated inception network for medical image segmentation.
\newblock \emph{international Journal of Advanced Computer Science and Applications}, 11\penalty0 (11), 2020.

\bibitem[Cao et~al.(2023)Cao, Wang, Chen, Jiang, Zhang, Tian, and Wang]{cao2023swin}
Hu~Cao, Yueyue Wang, Joy Chen, Dongsheng Jiang, Xiaopeng Zhang, Qi~Tian, and Manning Wang.
\newblock Swin-unet: Unet-like pure transformer for medical image segmentation.
\newblock In \emph{Computer Vision--ECCV 2022 Workshops: Tel Aviv, Israel, October 23--27, 2022, Proceedings, Part III}, pages 205--218. Springer, 2023.

\bibitem[Chen et~al.(2021)Chen, Lu, Yu, Luo, Adeli, Wang, Lu, Yuille, and Zhou]{chen2021transunet}
Jieneng Chen, Yongyi Lu, Qihang Yu, Xiangde Luo, Ehsan Adeli, Yan Wang, Le~Lu, Alan~L Yuille, and Yuyin Zhou.
\newblock Transunet: Transformers make strong encoders for medical image segmentation.
\newblock \emph{arXiv preprint arXiv:2102.04306}, 2021.

\bibitem[Codella et~al.(2019)]{codella2019skin}
Noel Codella et~al.
\newblock Skin lesion analysis toward melanoma detection 2018: A challenge hosted by the international skin imaging collaboration (isic).
\newblock \emph{arXiv preprint arXiv:1902.03368}, 2019.

\bibitem[Codella et~al.(2018)]{codella2018skin}
Noel~CF Codella et~al.
\newblock Skin lesion analysis toward melanoma detection: A challenge at the 2017 international symposium on biomedical imaging (isbi), hosted by the international skin imaging collaboration (isic).
\newblock In \emph{2018 IEEE 15th international symposium on biomedical imaging (ISBI 2018)}, pages 168--172. IEEE, 2018.

\bibitem[Deng et~al.(2009)Deng, Dong, Socher, Li, Li, and Fei-Fei]{deng2009imagenet}
Jia Deng, Wei Dong, Richard Socher, Li-Jia Li, Kai Li, and Li~Fei-Fei.
\newblock Imagenet: A large-scale hierarchical image database.
\newblock In \emph{2009 IEEE conference on computer vision and pattern recognition}, pages 248--255. Ieee, 2009.

\bibitem[Fiaz et~al.(2023)Fiaz, Heidari, Anwar, and Cholakkal]{fiaz2023sa2}
Mustansar Fiaz, Moein Heidari, Rao~Muhammad Anwar, and Hisham Cholakkal.
\newblock Sa2-net: Scale-aware attention network for microscopic image segmentation.
\newblock \emph{arXiv preprint arXiv:2309.16661}, 2023.

\bibitem[Fiaz et~al.(2024)Fiaz, Noman, and et~al.]{fiaz2024guided}
Mustansar Fiaz, Mubashir Noman, and et~al.
\newblock Guided-attention and gated-aggregation network for medical image segmentation.
\newblock \emph{Pattern Recognition}, page 110812, 2024.

\bibitem[Gu and Dao(2023)]{gu2023mamba}
Albert Gu and Tri Dao.
\newblock Mamba: Linear-time sequence modeling with selective state spaces.
\newblock \emph{arXiv preprint arXiv:2312.00752}, 2023.

\bibitem[Gupta et~al.(2021)Gupta, Gupta, Gehlot, and Goswami]{gupta2021segpc}
Anubha Gupta, Ritu Gupta, Shiv Gehlot, and Shubham Goswami.
\newblock Segpc-2021: Segmentation of multiple myeloma plasma cells in microscopic images.
\newblock \emph{IEEE Dataport}, 1\penalty0 (1):\penalty0 1, 2021.

\bibitem[He et~al.(2016)He, Zhang, Ren, and Sun]{he2016deep}
Kaiming He, Xiangyu Zhang, Shaoqing Ren, and Jian Sun.
\newblock Deep residual learning for image recognition.
\newblock In \emph{Proceedings of the IEEE conference on computer vision and pattern recognition}, pages 770--778, 2016.

\bibitem[Huang et~al.(2021)Huang, Deng, Li, and Yuan]{huang2021missformer}
Xiaohong Huang, Zhifang Deng, Dandan Li, and Xueguang Yuan.
\newblock Missformer: An effective medical image segmentation transformer.
\newblock \emph{arXiv preprint arXiv:2109.07162}, 2021.

\bibitem[Ibtehaz and Rahman(2020)]{ibtehaz2020multiresunet}
Nabil Ibtehaz and M~Sohel Rahman.
\newblock Multiresunet: Rethinking the u-net architecture for multimodal biomedical image segmentation.
\newblock \emph{Neural networks}, 121:\penalty0 74--87, 2020.

\bibitem[Kareem et~al.(2024)Kareem, Fiaz, Novershtern, and Cholakkal]{kareem2024medical}
Daniya Najiha~Abdul Kareem, Mustansar Fiaz, Noa Novershtern, and Hisham Cholakkal.
\newblock Medical image segmentation using directional window attention.
\newblock \emph{arXiv preprint arXiv:2406.17471}, 2024.

\bibitem[Leygeber et~al.(2019)Leygeber, Lindemann, Sachs, Kaganovitch, Wiechert, N{\"o}h, and Kohlheyer]{leygeber2019analyzing}
Markus Leygeber, Dorina Lindemann, Christian~Carsten Sachs, Eugen Kaganovitch, Wolfgang Wiechert, Katharina N{\"o}h, and Dietrich Kohlheyer.
\newblock Analyzing microbial population heterogeneity—expanding the toolbox of microfluidic single-cell cultivations.
\newblock \emph{Journal of molecular biology}, 431\penalty0 (23):\penalty0 4569--4588, 2019.

\bibitem[Lin et~al.(2022)Lin, Chen, Xu, Zhang, Lu, and Zhang]{lin2022ds}
Ailiang Lin, Bingzhi Chen, Jiayu Xu, Zheng Zhang, Guangming Lu, and David Zhang.
\newblock Ds-transunet: Dual swin transformer u-net for medical image segmentation.
\newblock \emph{IEEE Transactions on Instrumentation and Measurement}, 71:\penalty0 1--15, 2022.

\bibitem[Liu et~al.(2024)Liu, Yang, Zhou, Xi, Yu, Yu, Liang, Shi, Zhang, Zheng, and Wang]{SwinUMamba}
Jiarun Liu, Hao Yang, Hong-Yu Zhou, Yan Xi, Lequan Yu, Yizhou Yu, Yong Liang, Guangming Shi, Shaoting Zhang, Hairong Zheng, and Shanshan Wang.
\newblock Swin-umamba: Mamba-based unet with imagenet-based pretraining.
\newblock \emph{arXiv preprint arXiv:2402.03302}, 2024.

\bibitem[Ma et~al.(2024)Ma, Li, and Wang]{UMamba}
Jun Ma, Feifei Li, and Bo~Wang.
\newblock U-mamba: Enhancing long-range dependency for biomedical image segmentation.
\newblock \emph{arXiv preprint arXiv:2401.04722}, 2024.

\bibitem[Oktay et~al.(2018)Oktay, Schlemper, Folgoc, Lee, Heinrich, Misawa, Mori, McDonagh, Hammerla, Kainz, et~al.]{oktay2018attention}
Ozan Oktay, Jo~Schlemper, Loic~Le Folgoc, Matthew Lee, Mattias Heinrich, Kazunari Misawa, Kensaku Mori, Steven McDonagh, Nils~Y Hammerla, Bernhard Kainz, et~al.
\newblock Attention u-net: Learning where to look for the pancreas.
\newblock \emph{arXiv preprint arXiv:1804.03999}, 2018.

\bibitem[Ronneberger et~al.(2015)Ronneberger, Fischer, and Brox]{ronneberger2015u}
Olaf Ronneberger, Philipp Fischer, and Thomas Brox.
\newblock U-net: Convolutional networks for biomedical image segmentation.
\newblock In \emph{Medical Image Computing and Computer-Assisted Intervention--MICCAI 2015: 18th International Conference, Munich, Germany, October 5-9, 2015, Proceedings, Part III 18}, pages 234--241. Springer, 2015.

\bibitem[Sirinukunwattana et~al.(2017)Sirinukunwattana, Pluim, Chen, Qi, Heng, Guo, Wang, Matuszewski, Bruni, Sanchez, et~al.]{sirinukunwattana2017gland}
Korsuk Sirinukunwattana, Josien~PW Pluim, Hao Chen, Xiaojuan Qi, Pheng-Ann Heng, Yun~Bo Guo, Li~Yang Wang, Bogdan~J Matuszewski, Elia Bruni, Urko Sanchez, et~al.
\newblock Gland segmentation in colon histology images: The glas challenge contest.
\newblock \emph{Medical image analysis}, 35:\penalty0 489--502, 2017.

\bibitem[Szegedy et~al.(2015)Szegedy, Liu, Jia, Sermanet, Reed, Anguelov, Erhan, Vanhoucke, and Rabinovich]{szegedy2015going}
Christian Szegedy, Wei Liu, Yangqing Jia, Pierre Sermanet, Scott Reed, Dragomir Anguelov, Dumitru Erhan, Vincent Vanhoucke, and Andrew Rabinovich.
\newblock Going deeper with convolutions.
\newblock In \emph{Proceedings of the IEEE conference on computer vision and pattern recognition}, pages 1--9, 2015.

\bibitem[Szegedy et~al.(2016)Szegedy, Vanhoucke, Ioffe, Shlens, and Wojna]{szegedy2016rethinking}
Christian Szegedy, Vincent Vanhoucke, Sergey Ioffe, Jon Shlens, and Zbigniew Wojna.
\newblock Rethinking the inception architecture for computer vision.
\newblock In \emph{Proceedings of the IEEE conference on computer vision and pattern recognition}, pages 2818--2826, 2016.

\bibitem[Tragakis et~al.(2023)Tragakis, Kaul, Murray-Smith, and Husmeier]{tragakis2023fully}
Athanasios Tragakis, Chaitanya Kaul, Roderick Murray-Smith, and Dirk Husmeier.
\newblock The fully convolutional transformer for medical image segmentation.
\newblock In \emph{Proceedings of the IEEE/CVF Winter Conference on Applications of Computer Vision}, pages 3660--3669, 2023.

\bibitem[Valanarasu et~al.(2021)Valanarasu, Oza, Hacihaliloglu, and Patel]{valanarasu2021medical}
Jeya Maria~Jose Valanarasu, Poojan Oza, Ilker Hacihaliloglu, and Vishal~M Patel.
\newblock Medical transformer: Gated axial-attention for medical image segmentation.
\newblock In \emph{Medical Image Computing and Computer Assisted Intervention--MICCAI 2021: 24th International Conference, Strasbourg, France, September 27--October 1, 2021, Proceedings, Part I 24}, pages 36--46. Springer, 2021.

\bibitem[Wang et~al.(2022{\natexlab{a}})Wang, Cao, Wang, and Zaiane]{wang2022uctransnet}
Haonan Wang, Peng Cao, Jiaqi Wang, and Osmar~R Zaiane.
\newblock Uctransnet: rethinking the skip connections in u-net from a channel-wise perspective with transformer.
\newblock In \emph{Proceedings of the AAAI conference on artificial intelligence}, volume~36, pages 2441--2449, 2022{\natexlab{a}}.

\bibitem[Wang et~al.(2024{\natexlab{a}})Wang, Cao, Yang, and Zaiane]{wang2024narrowing}
Haonan Wang, Peng Cao, Jinzhu Yang, and Osmar Zaiane.
\newblock Narrowing the semantic gaps in u-net with learnable skip connections: The case of medical image segmentation.
\newblock \emph{Neural Networks}, page 106546, 2024{\natexlab{a}}.

\bibitem[Wang et~al.(2024{\natexlab{b}})Wang, Chen, Chen, and Wu]{lkmunet}
Jinhong Wang, Jintai Chen, Danny Chen, and Jian Wu.
\newblock Large kernel vision mamba unet for medical image segmentation.
\newblock \emph{arXiv preprint arXiv:2403.07332}, 2024{\natexlab{b}}.

\bibitem[Wang et~al.(2020)Wang, Hu, Cheah, Wang, Wang, Chen, Baikpour, Ozturk, Li, Chou, et~al.]{wang2020u}
Shuhang Wang, Szu-Yeu Hu, Eugene Cheah, Xiaohong Wang, Jingchao Wang, Lei Chen, Masoud Baikpour, Arinc Ozturk, Qian Li, Shinn-Huey Chou, et~al.
\newblock U-net using stacked dilated convolutions for medical image segmentation.
\newblock \emph{arXiv preprint arXiv:2004.03466}, 2020.

\bibitem[Wang et~al.(2022{\natexlab{b}})Wang, Xie, Li, Fan, Song, Liang, Lu, Luo, and Shao]{wang2022pvt}
Wenhai Wang, Enze Xie, Xiang Li, Deng-Ping Fan, Kaitao Song, Ding Liang, Tong Lu, Ping Luo, and Ling Shao.
\newblock Pvt v2: Improved baselines with pyramid vision transformer.
\newblock \emph{Computational Visual Media}, 8\penalty0 (3):\penalty0 415--424, 2022{\natexlab{b}}.

\bibitem[Xing et~al.(2024)Xing, Ye, Yang, Liu, and Zhu]{Segmamba}
Zhaohu Xing, Tian Ye, Yijun Yang, Guang Liu, and Lei Zhu.
\newblock { SegMamba: Long-range Sequential Modeling Mamba For 3D Medical Image Segmentation }.
\newblock In \emph{proceedings of Medical Image Computing and Computer Assisted Intervention -- MICCAI 2024}, volume LNCS 15008. Springer Nature Switzerland, October 2024.

\bibitem[Xu and Duan(2021)]{xu2021automatic}
Qing Xu and Wenting Duan.
\newblock An automatic nuclei image segmentation based on multi-scale split-attention u-net.
\newblock In \emph{MICCAI Workshop on Computational Pathology}, pages 236--245. PMLR, 2021.

\bibitem[Yuan et~al.(2023)Yuan, Zhang, and Fang]{yuan2023effective}
Feiniu Yuan, Zhengxiao Zhang, and Zhijun Fang.
\newblock An effective cnn and transformer complementary network for medical image segmentation.
\newblock \emph{Pattern Recognition}, 136:\penalty0 109228, 2023.

\bibitem[Zhou et~al.(2018)Zhou, Rahman~Siddiquee, Tajbakhsh, and Liang]{zhou2018unet++}
Zongwei Zhou, Md~Mahfuzur Rahman~Siddiquee, Nima Tajbakhsh, and Jianming Liang.
\newblock Unet++: A nested u-net architecture for medical image segmentation.
\newblock In \emph{Deep Learning in Medical Image Analysis and Multimodal Learning for Clinical Decision Support: 4th International Workshop, DLMIA 2018, and 8th International Workshop, ML-CDS 2018, Held in Conjunction with MICCAI 2018, Granada, Spain, September 20, 2018, Proceedings 4}, pages 3--11. Springer, 2018.

\end{thebibliography}
\end{document}